\documentclass[graybox]{svmult}

\usepackage{mathptmx}       
\usepackage{helvet}         
\usepackage{courier}        
\usepackage{type1cm}        
\usepackage{makeidx}         
\usepackage{graphicx}        
\usepackage{multicol}        
\usepackage[bottom]{footmisc}

\usepackage{amsmath}
\usepackage{amssymb}
\usepackage{algorithm}
\usepackage{algorithmic}
\usepackage{subfigure}
\usepackage{alltt}

\makeindex             

\begin{document}

\title*{Evolutionary Approaches to Explainable Machine Learning}
\author{Ryan Zhou and Ting Hu}
\institute{School of Computing, Queen's University, Kingston, ON K7L 2N8, Canada}

\maketitle

\abstract{
Machine learning models are increasingly being used in critical sectors, but their black-box nature has raised concerns about accountability and trust. The field of explainable artificial intelligence (XAI) or explainable machine learning (XML) has emerged in response to the need for human understanding of these models. Evolutionary computing, as a family of powerful optimization and learning tools, has significant potential to contribute to XAI/XML. In this chapter, we provide a brief introduction to XAI/XML and review various techniques in current use for explaining machine learning models. We then focus on how evolutionary computing can be used in XAI/XML, and review some approaches which incorporate EC techniques. We also discuss some open challenges in XAI/XML and opportunities for future research in this field using EC. Our aim is to demonstrate that evolutionary computing is well-suited for addressing current problems in explainability, and to encourage further exploration of these methods to contribute to the development of more transparent, trustworthy and accountable machine learning models.
}

\section{Introduction}

As the use of machine learning models becomes increasingly widespread in various domains, 
including critical sectors of society such as finance and healthcare, 
there is a growing need for human understanding of such models.
Although machine learning can detect complex patterns and relationships in data, 
decisions based on the predictions made by these models can have real-world impacts on human lives. 
The use of machine learning models in high-stakes applications such as medicine, job hiring, 
and criminal justice has raised concerns about the fairness, transparency, 
and accountability of these models. 
Therefore, it is essential not only to develop accurate prediction models 
but also to understand and explain how these predictions are being made.

The field of explainable artificial intelligence (XAI) or explainable machine learning (XML)  
has emerged in response to this need~\cite{iML_Murdoch_PNAS_19}. 
XAI research aims to develop methods to explain the decisions, predictions, 
or recommendations made by machine learning models in a way that is understandable to humans. 
These explanations act as a foundation to build trust and improve the robustness of a model 
by highlighting biases and failures, 
allow researchers to better understand, validate and debug the model, 
ensure compliance with regulations, 
and improve human-machine interaction by giving users a better understanding of 
when they can rely on a model's decisions.

In this book chapter, 
we aim to provide an overview of XAI/XML 
and the role of evolutionary computing (EC) in this field. 
In Section~\ref{section:xai}, 
we introduce the concept of XML and review current techniques used in the field. 
In Section~\ref{section:ec_for_xai}, 
we will focus on evolutionary techniques for explaining machine learning models. 
Specifically, we will cover data visualization, feature selection and engineering, 
local and global explanations, counter-factual and adversarial examples. 
We will also discuss evolutionary methods designed specifically for explaining deep learning models, 
as well as evolutionary methods for assessing the quality of explanations themselves. 
In Section~\ref{section:outlook}, 
we identify some challenges remaining in the field of XAI/XML 
and discuss future opportunities for incorporating EC. 
Finally, Section~\ref{section:conclusion} provides concluding remarks for the chapter.

\section{Explainable Machine Learning}
\label{section:xai}

Explainability and the fields of XAI/XML are concerned with 
extracting insights from machine learning models in order to 
make them more transparent, interpretable and understandable to humans. 
The goal of XAI/XML is to develop methods that 
can provide clear explanations of the decision-making process and enable humans to understand 
what relationships a model has learned and how the model arrived at its conclusions.

More concretely, the types of questions we wish to answer with an explanation 
include~\cite{bacarditIntersectionEvolutionaryComputation2022a}:
\begin{itemize}
\item Are the patterns the model is drawing on to make its prediction the ones we expect?
\item Why did the model make this prediction instead of a different one, 
and what would it take to make it change its prediction?
\item Is the model biased and are the decisions made by the model fair?
\end{itemize}

Explanations can be provided in many forms; 
examples include visualizations, numerical values, data instances, 
or text explanations \cite{iML_Murdoch_PNAS_19}.

\subsection{Interpretability vs.~Explainability}

The terms interpretability and explainability are often used interchangeably by researchers. 
However, in this chapter we distinguish between them as referring to two different 
but related aspects of attempting 
to understand a model~\cite{liptonMythosModelInterpretability2018,iML_Rudin_NatML_19}.

For our purposes, 
interpretability refers to the level of transparency and ease of understanding 
of a model's decision-making process. 
An \textit{interpretable} model is one whose decisions can be traced and understood by a human, 
for example, by having a simple symbolic representation. 
In other words, a model is considered interpretable 
if its decision-making process can be understood simply by inspecting the model. 
For example, small decision trees are often considered interpretable 
because it is feasible for a human to follow the exact steps leading to the prediction.

On the other hand, even if we cannot trace the exact logic, 
a model can still be considered \textit{explainable} 
if a human-understandable explanation can be provided for what the model is doing 
or why a decision is made. 
The more comprehensive and understandable the explanation is, 
the more explainable the overall system is. 
Explanations can be provided for the overall patterns and relationships learned by a model, 
or for the logic used to produce an individual prediction. 
Some practical methods for providing explanations include evaluating feature importance, 
visualizing intermediate representations, or comparing the prediction to be explained with samples from the training set. An explanation essentially acts as an interface between the human and the model, attempting to approximate the behavior or decisions made by the model in a way 
that is comprehensible to humans. As such, this ``model of the model" must be both an accurate proxy for the model in the areas of interest but also be understandable itself.

These two terms, interpretability and explainability, also roughly correspond to two paradigms for improving the understandability of a model. The first approach, often known as ``intrinsic interpretability", ``glass-box" or ``interpretability by design", focuses on the interpretability of the model and aims to improve it by simplifying or otherwise restricting its complexity~\cite{IML_eBook_Molnar_22}. 
A simple enough model allows the exact logic to be easily followed and understood by humans, 
but over-reducing the model complexity can also restrict the performance of the model. 
In some cases, it might not be possible to construct a simple enough model 
that can still effectively solve the problem at hand.

The second approach, known as ``post-hoc" or ``black-box" methods, 
focuses on explainability and aims to provide explanations by analyzing the model after it has been trained. This approach does not restrict the performance of the model in any way, 
but also makes it more challenging to provide informative explanations. 
These explanation methods are typically independent of the original model 
and are learned after the initial model has been trained~\cite{IML_eBook_Molnar_22}. 
In this chapter, we will primarily focus on this approach to understanding models. 
However, in some cases, an explanation for a complex model can take the form of simpler model; 
in this case, it is important to consider the interpretability of the explanation as well.

It is also worth noting that these two goals are not mutually exclusive, and both approaches can be used to better understand a model. The overall goal of explainability is to develop better tools to understand machine learning models, while the goal of interpretability is to construct models that can be understood using the tools we have.

\subsection{Explanation Methods}

\begin{figure}[t!]
  \includegraphics[width=\linewidth]{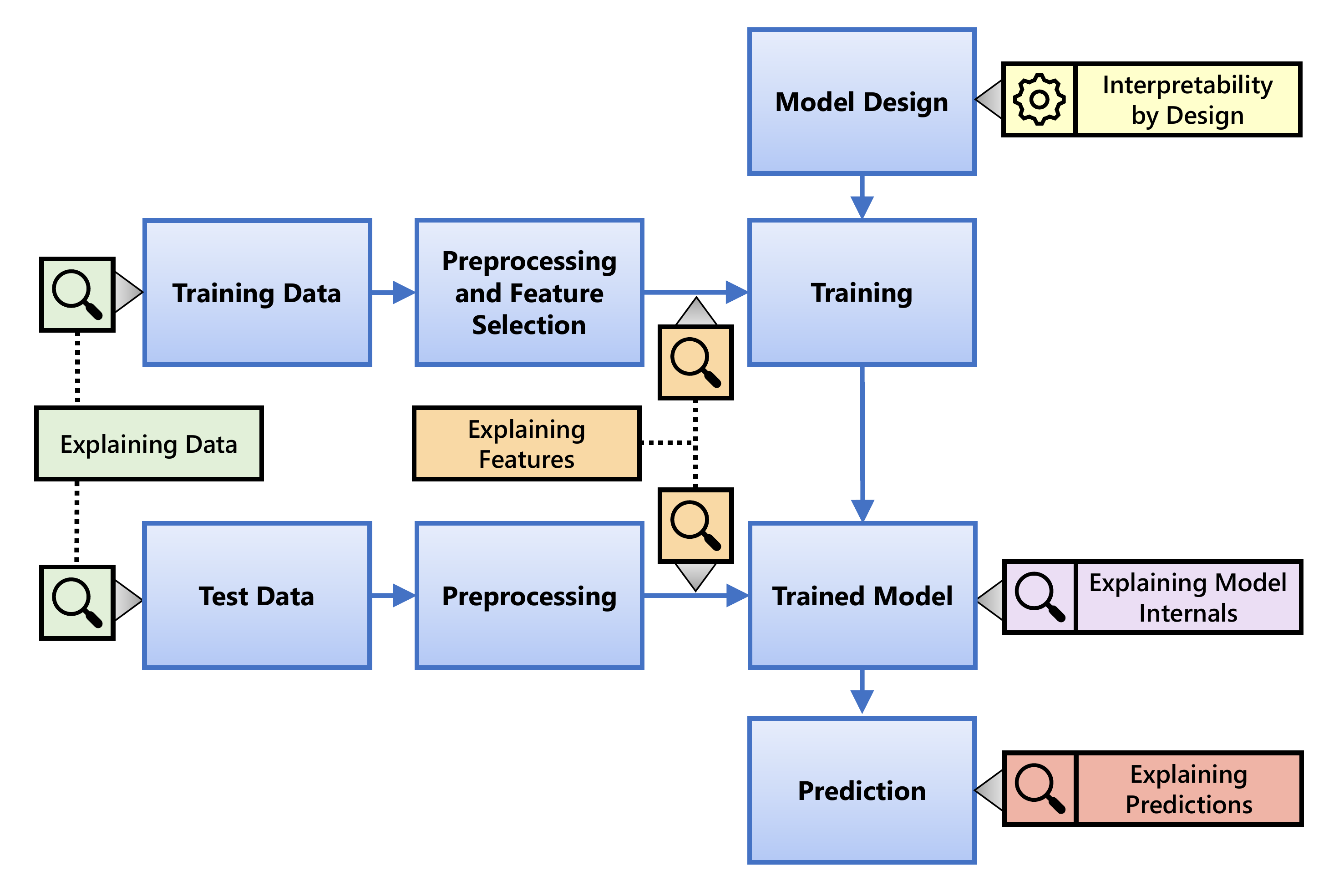}
  \caption{Overview of the process of building a machine learning model, 
  showing areas where explanations (magnifying glasses) are often applied. 
  Examples of methods in each category are described in Section~\ref{section:xai}. 
  Also shown is the intrinsic interpretability approach (cogwheel), 
  where models are designed to be interpretable from the start. 
  All these methods can be used together to form a more complete picture of a model's behavior.}
  \label{fig:xai}
\end{figure}

There are several types of explanations, 
which focus on different aspects of the modeling process (Figure~\ref{fig:xai}). 
In this section, we will provide an overview of various categories, 
and describe some examples of each. 
This overview is not meant to be exhaustive, 
but rather to highlight the various areas where XAI techniques can be applied.
For a more intensive survey of current methods in XAI, we direct the reader to recent reviews \cite{dwivediExplainableAIXAI2023, saeedExplainableAIXAI2023} on the topic.

\subsubsection{Explaining Data}
Although this category is sometimes omitted in discussions of explainable machine learning, 
it is worth mentioning as part of the overall pipeline. Methods under this category do not necessarily explain the model itself, but aim to explain the underlying data that a model is trained on, focusing on understanding the data distribution and its characteristics. Techniques such as exploratory analysis, data visualization and dimensionality reduction can be used to gain a better understanding of the patterns in the underlying data that the model might learn, as well as identify any potential biases. Examples of these techniques include Principal Component Analysis (PCA)~\cite{pearsonLIIILinesPlanes1901, hotellingAnalysisComplexStatistical1933} and t-Distributed Stochastic Neighbor Embedding (t-SNE)~\cite{maatenVisualizingDataUsing2008}, which reduce the dimensionality of data to allow for easy visualization. In addition, methods such as clustering and outlier detection can help identify patterns or anomalies in the data that may impact the model's performance, and can aid in feature selection and engineering. These explanations can help identify data quality issues, biases, and preprocessing requirements, as well as build trust.


\subsubsection{Explaining Features}
This approach aims to explain the dependence of a model on each feature it uses. 
For example, feature importance returns a score 
that represents the significance of each feature to the model. 
This helps to identify which features have the greatest impact on the model's predictions 
and provides insights into how the model is making decisions. 
This type of explanation can also be used to verify whether the model is behaving as expected 
- for example, by checking whether it is using the same features a human would to solve the problem. 
In the case of a computer vision model, 
this type of explanation can be used to determine 
if the features being used to classify a particular image as a cat make sense, 
or if the model is using spurious patterns in the data, 
such as identifying the cat based on its surroundings. 
This type of explanation can also aid in optimizing models 
and performing feature selection by identifying less important features.

Some models, such as decision trees and tree ensembles like random forests, 
provide feature importance measures~\cite{randForest_Breiman_ML_01}. 
For other models, Shapley additive explanations (SHAP)~\cite{SHAP_Lundberg_NeurIPS_17} 
attempts to provide a universal method for assessing feature importance 
that can be applied to most machine learning models. 
This is based on the Shapley value, 
a concept from cooperative game theory that assigns a value to each player in a game 
based on their contribution to the overall outcome. 
In the context of machine learning, 
the ``players" are the features in the data, and the ``game" is the prediction task. 
The Shapley value scores each feature based on its contribution to each prediction. 
The exact calculation of Shapley values is usually computationally impractical, 
as it involves evaluating every possible combination of features. 
However, SHAP proposes approximating these values using sampling and regression, 
making the estimation of feature importance computationally feasible. 
This method is widely used in the field of XAI.

\subsubsection{Explaining Model Internals}

Methods in this category attempt to explain the internal function of the model, 
for example, by inspecting the structure of a tree or the weights in a neural network. 
These approaches can either attempt to explain the entire model, 
or to understand smaller components of the model.

Explaining the full model can be done by training a secondary model 
which both approximates the original model and is more interpretable. 
An example of this approach was proposed by 
Lakkaraju et al.~\cite{lakkarajuInterpretableExplorableApproximations2017}. 
Their approach approximates the behavior of the model using a small number of decision sets, 
providing an interpretable proxy for the entire model. 
While approximating the full model in this and similar ways is effective for smaller models, 
producing accurate but interpretable proxy models 
becomes increasingly difficult as the size of the model increases. 
For large models with sufficiently complex behavior, 
the proxy model will either be a poor approximation 
or will itself be so large that it becomes difficult to interpret.

This difficulty has led to the development of methods 
which attempt to explain smaller sub-components of a large model. 
Examining sub-components can allow us to break down the overall function in a modular way, 
or to identify the parts of a model which are responsible for certain decisions. 
For example, one way to explain the function of a neuron in a neural network is 
by finding or constructing an input that will maximize 
its activation~\cite{mahendranVisualizingDeepConvolutional2016}. 
This produces an idealized version of the input that the neuron activates on.
 
In recent years, many methods have been developed for explaining 
the internals of deep learning models~\cite{samekExplainingDeepNeural2021}. 
This is due to the rise in popularity of such models and their inherent black-box nature and large size, 
making explanations for them challenging but particularly important. 
As an example of one such method, 
Interpretable Lens Variable Models~\cite{adelDiscoveringInterpretableRepresentations2018} 
train an invertible mapping from a complex internal representation inside a neural network 
(the latent space in a generative or discriminative model) to a simpler, interpretable one. 
A user supplies some side information in the form of a few examples which illustrate the properties they are interested in being represented explicitly, such as rotation and scaling. Then, the mapping is learned while constraining certain dimensions in the interpretable representation to obey a linear relationship with the side information. Since the mapping is invertible, the output of the model can also be controlled by making changes to the interpretable representation. The authors' experiments showed that this method can enable the user to control these properties of the images generated by the model in an intuitive way.

\subsubsection{Explaining Predictions}

This type of approach aims to explain a specific prediction made by a model. 
As such, the explanation only needs to capture the behaviour of the model 
with respect to the prediction in question, rather than the model as a whole.

One popular approach in this category is 
Local Interpretable Model-agnostic Explanations (LIME)~\cite{expPredict_Ribeiro_KDD_16}. 
LIME explains a prediction by sampling a set of instances similar to the input to be explained. 
It obtains predictions for each of these instances and then fits a linear model to the sampled set of inputs and predictions. This creates a linear approximation for the model 
in the local neighborhood surrounding the instance to be explained. 
Although this explanation does not necessarily reflect the global behavior of the model, 
it is locally faithful and allows us to understand the behavior of the model around that point.

Counterfactual explanations are another type of explanation 
which provide information through a hypothetical example 
in which the model would have made a different decision. 
For example, ``the model would have approved a loan if your income were \$5000 higher" 
is a counterfactual illustrating how the input would need to be changed 
in order to get a different result from the model~\cite{wachterCounterfactualExplanationsOpening2017b}. 
This type of explanation is intuitive and can be performed on a model in a black-box manner, 
without any access to a model's internals. 
One advantage of counterfactual explanations is that 
they can provide users with \textit{recourse}, 
a concrete set of changes that could be made to change the decision to 
a different one~\cite{karimiSurveyAlgorithmicRecourse2020}. 
Additionally, because they directly operate on the model's evaluation, 
they are always faithful to the model's true behavior. 
However, because they consist of single instances or data points, 
they provide less insight into the model's overall behavior 
compared to a more comprehensive explanation.

Diverse Counterfactual Explanations (DiCE)~\cite{mothilalExplainingMachineLearning2020} 
is one common method of constructing counterfactual predictions. 
The aim of this method is to produce counterfactuals 
which are valid (produce a different result when fed into the model), 
proximal (are similar to the input), 
and diverse (different from each other). 
Diversity is desirable here as it increases the likelihood of 
finding a useful explanation and provides a more complete picture of the model's behavior. 
DiCE generates a diverse set of counterfactual examples using a diversity metric 
based on determinantal point processes~\cite{kuleszaDeterminantalPointProcesses2012}, 
a probabilistic model which can solve subset selection problems under diversity constraints. 
This diversity constraint forces the various examples apart, 
while an additional proximity constraint forces the examples to lie close to the original input. 
The method also attempts to make the counterfactual examples differ from the input 
in as few features as possible (feature sparsity).

\subsubsection{Other Considerations}

When choosing an explanation method, there are some other factors to consider, 
such as whether a method is {\it model-specific} or {\it model-agnostic}, 
and whether it provides {\it global} or {\it local} explanations.

Whether an explanation is considered model-specific or model-agnostic 
is based on whether the technique is restricted to certain types of models, 
or if it can be applied to multiple types of models. 
For instance, some methods of generating feature importance measures~\cite{randForest_Breiman_ML_01} 
are specific to random forests and cannot be directly used to 
generate feature importance for a neural network. 
Conversely, some methods are designed for extracting information 
from the internal representation of neural networks, 
and cannot be applied to random forests. 
These methods are model-specific. 
On the other hand, Shapley additive explanations (SHAP) can be applied to 
assess feature importance for any of the commonly used machine learning models, 
making it model-agnostic.

In addition, explanations can be local or global. 
Local explanations aim to provide an explanation for a specific input or group of inputs, 
while global explanations aim to explain the behavior of the entire model. 
For instance, a global explanation technique such as 
partial dependence plots~\cite{EstatLearn_Hastie_01} or feature importance 
identifies influential features across the entire training dataset or model. 
On the other hand, LIME fits its linear model to the neighborhood of a specific prediction, 
making it a local explanation as it only deals with the behavior of the model 
in the specific region near the prediction. 
While global explanations may appear to be preferable to local explanations, 
in practice explanations are always constrained by the amount of information 
a human can grasp at one time. 
As such, while global explanations can provide information 
about overall trends in the data or model, 
local explanations are useful for providing more detail about specific aspects of interest.

\section{Evolutionary Approaches to Explainable Machine Learning}
\label{section:ec_for_xai}

\subsection{Why Use EC?}

Evolutionary computing (EC) is an approach to automatic adaptation inspired 
by the principles of biological evolution that has been successfully applied to various fields, 
including optimization, learning, engineering design, and artificial life. 
This approach mimics the process of natural selection, 
allowing the computer to evolve and optimize solutions to complex problems 
by combining and improving upon existing solutions.
Common techniques within EC include evolutionary algorithms such as genetic algorithms, genetic programming, and evolution strategies, 
and swarm intelligence algorithms such as particle swarm optimization. 
These techniques use iterative processes of selection, recombination, mutation, 
and adaptation to improve the performance of the models.

EC has unique strengths 
that make it well-suited for handing the challenges of XAI. 
First of all, EC can work with symbolic representations 
or interpretable models such as decision trees or rule systems. 
This can be used to guarantee interpretability of the evolved representation. 
If EC is used to evolve an explanation using interpretable components, 
the explanation will be interpretable. 
To leverage this, a common approach is to evolve an interpretable approximation of the model, 
in whole or in part.

Evolutionary methods are also highly flexible, 
being able to perform black-box optimization, 
and usually being derivative-free. 
As a result, they can be applied to a variety of models without requiring knowledge of the model's internal logic -- 
for example, when access to a model is only available through an API which returns only the predictions. 
This flexibility also enables optimization of unusual or customized metrics without requiring in-depth knowledge of how to optimize the metric. 
Additionally, the flexibility of evolutionary methods allows them
to be used to create hybrid methods with other algorithms 
or to serve as meta-learning methods.

EC is advantageous for XAI as it can handle multiple objectives simultaneously. 
This is important as explanations by nature require both faithfulness to the model 
as well as interpretability to humans. 
The population-based nature of evolutionary algorithms also enables explicit use of diversity metrics. 
This allows us to explore a range of different explanations, increasing the chances of finding a useful one.

We will now delve into current methods which employ EC for producing explanations, 
and provide an overview of these approaches and their applications. 
There are a number of different ways to generate explanations, and as before, 
we will discuss them based on the stage in the machine learning modeling process in which they can be applied, highlighting works of interest. For an extensive survey on these methods, we invite the reader to explore recent reviews~\cite{bacarditIntersectionEvolutionaryComputation2022a, meiExplainableArtificialIntelligence2022} on the use of evolutionary methods in XAI.

\subsection{Explaining Data}

EC can be used to explain data by means of dimensionality reduction and visualization. One approach is GP-tSNE~\cite{GPinterpVis_Lensen_ITC_21}, which adapts the classic t-SNE~\cite{maatenVisualizingDataUsing2008} algorithm to use evolved trees to provide an interpretable mapping from the original data points to the embedded points. Similarly, Schofield and Lensen~\cite{schofieldUsingGeneticProgramming2021} use tree-GP to produce an interpretable mapping for Uniform Manifold Approximation and Projection (UMAP). By producing an explicit mapping function rather than simply the embedded points, we can not only make the process more transparent but also reuse the mapping on new data.

In some cases, we may want to use the lower dimensional representation for prediction as well as visualization. This is useful for interpretability as it allows us to visualize exactly the same data representation that the model sees. Therefore, another approach is to construct features which are both amenable to visualization and well-suited for downstream tasks. Icke and Rosenberg~\cite{ickeMultiobjectiveGeneticProgramming2011} proposed a multi-objective GP algorithm to optimize three objective measures desirable for constructed features - classifiability, visual interpretability and semantic interpretability. Similarly, Cano et al.~\cite{canoMultiobjectiveGeneticProgramming2017} developed a method using multi-objective GP to construct features for visualization and downstream analysis, optimizing for six classification and visualization metrics. The classification metrics (accuracy, AUC and Cohen's kappa rate) aim to improve the performance of the downstream classifier, while the visualization metrics (C-index, Davies-Bouldin index, Dunn's index) aim to improve the clustering and separability of the features.

\subsection{Feature Selection and Feature Engineering}
Feature selection is a common preprocessing step where a relevant subset of features is selected from the original dataset. This is used to improve the performance of the model, but also has benefits for interpretability by narrowing down the features the model can draw on. As an explanation, feature selection shares some similarities with feature importance, which identifies the features a model is drawing on, but instead restricts the model explicitly so it can only draw on the chosen features.

Genetic algorithms are a straightforward and effective approach to feature selection, with a natural representation in the form of strings of 1s and 0s, making them a popular choice for feature selection~\cite{sayedNestedGeneticAlgorithm2019, shaFeatureSelectionPolygenic2021,xueMultiObjectiveFeatureSelection2022}. 
Genetic programming can also be used for feature selection since the inclusion of features in a tree or linear genetic program is intrinsically evolved with the program~\cite{GPinterpret_Hu_GPTP_20,EvoNetOA_Hu_PLOSCB_18,SMILE_Sha_BMCBio_21}.
For an in-depth review of genetic programming methods, we refer the reader to~\cite{xueSurveyEvolutionaryComputation2016}. Swarm intelligence methods, such as particle swarm optimization, have been applied to feature selection as well~\cite{xueSelfadaptiveParticleSwarm2019}. For a more detailed review of these methods, we direct the reader to~\cite{nguyenSurveySwarmIntelligence2020}. In addition to selecting features for a model, feature selection can also be used to improve data understanding by integrating it with techniques such as clustering ~\cite{hancerSurveyFeatureSelection2020}.


Feature engineering, also referred to as feature construction, is a related approach that involves building higher-level condensed features out of basic features. Genetic programming can be used to evolve these higher-level features for downstream tasks such as classification and regression~\cite{lacavaLearningFeatureSpaces2020,GP-FE_Li_BIBM_20, muharramEvolutionaryConstructiveInduction2005, virgolinExplainingMachineLearning2020}. This can also help improve the interpretability of a model, since this can reduce a large number of low-level features to a smaller number of higher-level features which may be easier to understand for humans. Moreover, it removes some of the modeling out of the black-box and into a transparent step beforehand, thereby reducing the amount of explanation needed.

These methods also share many similarities with dimensionality reduction techniques, and in some cases can fall under both categories. Uriot et al.~\cite{uriotGeneticProgrammingRepresentations2022} compared a variety of multi-tree GP algorithms for dimensionality reduction, as well as a tree-based autoencoder architecture. In the multi-tree representation, each individual in the population is a collection of trees with each tree mapping the input to one feature in the latent dimension. In order to reconstruct the input for the autoencoder, a multi-tree decoder is simultaneously evolved with one tree per input dimension. Their results showed that GP-based dimensionality reduction was on par with the conventional methods they tested (PCA, LLE, and Isomap).

\subsection{Model Extraction}

This approach, also known as a global surrogate model, aims to approximate a black-box model with an evolved interpretable model. This idea is closely related to knowledge distillation~\cite{buciluaModelCompression2006,hintonDistillingKnowledgeNeural2015} in deep learning, but rather than simply making the model smaller we also want to make it more interpretable.

Evans et al.~\cite{iMLGP_Evans_GECCO_19} propose a model extraction method using multi-objective genetic programming to construct decision trees that accurately represent any black-box classifier while being more interpretable. This method aims to simultaneously maximize the ability of the tree to reconstruct (replicate) the predictions of a black-box model and also maximize interpretability by minimizing the complexity of the decision tree. The reconstruction ability is measured by the weighted F1 score over cross-validation, and the complexity of the decision tree is measured by the number of splitting points in the tree. The overall evolutionary process uses a modified version of NSGA-II~\cite{NSGA-II_Deb_TEC_02}. In their experiments on a range of classification problems, they found that the accuracy remained commensurate with other model extraction methods (Bayesian rule lists, logistic regression, and two types of decision trees) while significantly reducing the complexity of the models produced.

\subsection{Local Explanations}

Instead of creating an interpretable model to approximate the global performance of a black-box model, which may not be possible, these approaches only attempt to approximate the local behaviour using an evolutionary algorithm. Much work in this area is inspired by LIME, but enhance the relatively simple sampling strategy and linear approximation with evolutionary components.

Ferreira et al.~\cite{GP-localExp_Ferreira_CEC_20} proposed Genetic Programming Explainer (GPX), a GP-based method which fits a local explanation model for a given input example. Similar to LIME, when given a sample input to be explained, the method samples a set of neighboring data points around the input and fits a local explanation. However, rather than a linear model, GPX uses a GP to evolve symbolic expression trees that best capture the behavior of the pre-trained black-box model over the neighboring data points. The authors tested this on both classification and regression datasets and reported that the GP captured the model's behavior better than LIME, as the assumption of linear local behavior was not always valid, and also outperformed a decision tree used as an explainer for the same neighbor set.

On the other hand, Guidotti et al.~\cite{guidottiLocalRuleBasedExplanations2018} proposed Local Rule-based Explanations (LORE), which applies an evolutionary algorithm to neighborhood generation rather than evolving the explanation itself. A genetic algorithm generates a set of points near the prediction to be explained, which are either classified the same as or differently from the original prediction while being nearby. A decision tree is then used to fit the local behavior of the black-box model. The use of a genetic algorithm here ensures a dense sampling of points in the local neighborhood which lie on both sides of the decision boundary.

\subsection{Counterfactuals}

Another line of work is in creating counterfactuals using EC. An EA is well suited to this task as a black-box, possibly multi-objective optimizer, as it allows us to find counterfactuals without knowing the internal workings of the model while also optimizing for multiple desirable criteria in the counterfactuals.

CERTIFAI~\cite{sharmaCERTIFAICommonFramework2020} generates a population of counterfactual explanations using a model-agnostic genetic algorithm. The initial population is generated by sampling instances which lie on the other side of the decision boundary of the model (i.e., are classified differently from the instance to be explained). Then, the genetic algorithm optimizes the population to minimize the distance (for some notion of distance, depending on the type of data) from each counterfactual instance to the input instance. The population is then analyzed for robustness, which increases if the best counterfactual examples found are farther away from the input, and fairness, which is measured by comparing robustness across different values of a particular feature.

GeCo~\cite{schleichGeCoQualityCounterfactual2021} uses a genetic algorithm with feasibility and plausibility constraints on the features, specified using the constraint language PLAF. This allows certain counterfactuals which would be useless to the user (e.g. counterfactuals where the user changes their gender or decreases their age) to be ruled out. Similar to CERTIFAI, the genetic algorithm minimizes the distance from the input instance to the counterfactual examples, prioritizing examples on the other side of the decision boundary but also keeping examples which are close to the decision boundary if not enough counterfactuals are available. The fitness function does not consider how many features are changed relative to the input instance (with a smaller number being preferred for ease of understanding), but the algorithm is biased toward a smaller number of changes by initializing the population with only one feature changed.

Multi-objective counterfactuals (MOC)~\cite{dandlMultiObjectiveCounterfactualExplanations2020} explicitly use multi-objective optimization to consider multiple desirable properties of the explanations. MOC uses a modified version of NSGA-II to perform its search. Among the changes are the use of mixed integer evolution strategies (MIES)~\cite{liMixedIntegerEvolution2013} to search a mixed discrete and continuous space, and a different crowding distance sorting algorithm which prioritizes diversity in feature space. A total of four objectives are used, optimizing for these four desirable properties: the model output for the example should be close to the desired output; the example should lie close in feature space to the input to be explained; the example should not differ from the input in too many features; and the example is plausible (likely to be drawn from the same distribution as the real data), measured by its distance to the closest $k$ data points.

\subsection{Explaining Deep Learning}

Thus far, the methods we have covered are general, and can be used with a variety of models. However, with the popularity of deep learning methods, we would be remiss not to discuss some specific methods tailored for deep learning models. While deep learning models are large, they are also differentiable allowing for hybrid methods combining gradient information with evolution.

For image classification, the large number of input features (pixels) presents a significant problem for many explanation methods. As such, it is necessary to reduce the dimension first, for example by clustering similar pixels into ``superpixels". 
Wang et al.~\cite{wangMultiobjectiveGeneticAlgorithm2022} propose using a multi-objective genetic algorithm to identify superpixels of importance for the final prediction, and using this set of superpixels as an explanation. The genetic algorithm uses NSGA-II to optimize for the least number of superpixels used, while maximizing the model's confidence in its prediction.

Adversarial examples are closely related to counterfactuals. An adversarial example is a counterfactual example but with the intent of creating an incorrect prediction~\cite{IML_eBook_Molnar_22}. This is done by applying a small perturbation to an example to change its classification. Most approaches search for examples which are as close to the original input as possible, and perceptually similar to the input. These examples are a method to highlight failure modes of the model as well as a potential attack vector on deep learning models.

Su et al.~\cite{suOnePixelAttack2019} propose a method of finding adversarial examples which modify only one pixel in an image. This is in contrast to previous methods which modify multiple pixels in the image and are more obvious to humans. Their method uses differential evolution, where each individual is encoded by the coordinate of the pixel to be modified and the perturbation in RGB space. They find that in many cases one pixel is sufficient to deceive the model.

Adversarial examples are also present in models built for other domains, such as natural language processing. Alzantot et al.~\cite{alzantotGeneratingNaturalLanguage2018} generate adversarial examples on a sentiment analysis model and a textual entailment model. In addition, the examples they produce are designed to be semantically and syntactically similar to the original input, making the attack more difficult to spot. A genetic algorithm is used to optimize for a different target label than the original. Mutation is done by changing words in the input to similar words as measured by a word embedding model (GloVe), and filtering out words which do not fit the context.

\subsection{Assessing Explanations}

Finally, rather than using EC to generate the explanations themselves, we will discuss some ways that it can be used to assess or improve the quality of other explanation methods.

Huang et al.~\cite{huangSAFARIVersatileEfficient2022} propose two metrics to assess the robustness of an explanation: worst-case misinterpretation discrepancy and probabilistic interpretation robustness. Interpretation discrepancy measures the difference between two interpretations, one before and one after perturbation of the input. It is desirable for this value to be low for a interpretation to be robust to perturbation. They then measure the discrepancy in two worst cases: the largest interpretation discrepancy possible while still being classified as the same class, and the smallest interpretation discrepancy possible while being classified differently (adversarial example). These values are optimized for using a GA. The other metric, probability of misinterpretation, calculates probabilistic versions of the above: the probability of an example having the same classification but significantly different interpretation, and the probability of an example having a different classification but similar interpretation. This is estimated using subset simulation.

It is also possible to perform an adversarial attack on the explanations themselves. Tamam et al.~\cite{tamamFoilingExplanationsDeep2022} do this with AttaXAI, a black-box approach based on  evolution. AttaXAI tries to evolve an image similar in appearance to the original input and producing the same prediction from the model, but with an arbitrary explanation map. In their experiments, pairs of images were selected and they showed that they were able to generate an new image with the appearance and prediction of the first image, but with a similar explanation map to the second.

\section{Research Outlook}
\label{section:outlook}

\subsection{Challenges}

One major challenge for evolutionary approaches to XAI is scalability. As data continues to grow and machine learning models become increasingly complex, the number of parameters and features to be optimized grows as well. As such, methods which work well on small models and datasets may become too expensive on larger ones. However, large models are the most opaque and most in need of explanation, so improving the scalability of XAI methods is necessary to ensure they can be applied to even the largest models. In particular, producing fully interpretable global explanations which accurately capture behavior yet are still simple enough to understand may become too challenging as models become larger -- necessitating more local explanations or a more focused approach concentrating on explaining particular properties or components of the model. We also see here the potential for more use of automated approaches to explainability -- for example, by using evolutionary search to find local explanations of interest and optimize for particular properties. This idea has been explored with counterfactual examples, but it could be extended to other types of explanations.

Another challenge for explainability is the incorporation of domain knowledge. This can include knowledge from subject matter experts, as well as prior knowledge about the dataset or problem. Current approaches to XAI are broad and aim to provide explanations which are independent of the problem setting, or at most are model-specific rather than problem-specific. However, it can useful to see how well a solution found by a machine learning model aligns with current knowledge in the field to evaluate the quality of the solution, or conversely, to identify areas where the model deviates from current understanding. For example, a practitioner may want to see how well the gene associations found by a genomics model aligns with the literature, as well as which associations are novel. This domain knowledge can be provided in the form of expert rules, constraints, or structured data such as a graph structure or tree. Domain knowledge can also be incorporated into the model-building process to improve interpretability, for instance by constraining the models to focus on associations known to be plausible (e.g. by incorporating causality) or excluding irrelevant features.

\subsection{Opportunities}

We see some additional opportunities for future work employing EC for XAI. One promising direction in current research is the use of multiple objectives to optimize explanations. Explainability is inherently a multi-objective problem, requiring the explanation to both be faithful to the model as well as being simple enough to be interpretable. EC is well-suited to explicitly optimizing for this, and we believe introducing these ideas into current and future explanation methods is a straightforward but effective way of improving the quality of explanations.

Along similar lines, diversity metrics and novelty search are another unique strength available to evolutionary algorithms which can help improve the explanations provided. The use of quality-diversity (illumination) algorithms can produce a range of explanations which are both accurate and present different perspectives on the behavior of the model. 
For example, a quality-diversity approach to counterfactual explanations could ensure that a range of behaviors are showcased in the examples.

Another opportunity for EC is the incorporation of user feedback, considering the evolution of explanations as an open-ended evolution process. Explainability is intended for the human user, and as such explanation quality is ultimately subjective and can only be approximated by metrics. Users may also have their own unique preferences for what constitutes a useful explanation. Incorporating user feedback into the evolution process can allow better tailored explanations that continue to improve. At the same time, better metrics for measuring the quality of an explanation are also necessary in order to not overwhelm the user.

\subsection{Real-world Impacts}

As AI becomes increasingly integrated into real-world applications, developing better methods for providing explanations is essential for ensuring safety and trust across various domains. With this in mind, it is also crucial to consider the practical effects and benefits that XAI research can have. We would like to highlight here a few application areas where work on evolutionary approaches to XAI can have a substantial impact.

Healthcare is a domain where the consequences of errors can be especially high. Untrusted models may be ignored by clinicians, wasting resources and providing no benefit. Worse, seemingly trustworthy but flawed models may cause harm to patients. Even models with few errors may exhibit systematic biases, such as diagnostic models underdiagnosing certain patient groups while appearing accurate~\cite{seyyed-kalantariUnderdiagnosisBiasArtificial2021}. Explainability can help identify these systematic errors and biases~\cite{arias-duartFocusRatingXAI2022}. In the financial sector AI models are employed for fraud detection and risk assessment. Similar systematic biases in these models can also produce harm, for example by disproportionately denying loans to certain groups. In addition, regulatory bodies often require explanations for these models to ensure compliance and maintain transparency.

Explainability also holds significant potential to advance engineering and scientific discovery. AI models are used in various engineering applications, for example in AI-driven materials design and drug discovery, and to produce scientific insights in fields such as genomics and astrophysics. Explanations can offer insight into the underlying mechanisms and relationships, improving hypothesis generation and validating domain knowledge.

Natural language processing has experienced many recent breakthroughs, with the development and deployment of models of unprecedented size. In particular, there is an emerging paradigm of building ``foundation models", generalist deep learning models which are trained on a wide range of data for general capabilities and which can be further fine-tuned for downstream tasks \cite{bommasaniOpportunitiesRisksFoundation2021}. These models are capable of tasks which they are not specifically trained for, but it is still unclear how they make decisions or generate outputs. Any flaws in these foundation models may be carried over to application-specific models built on top of them. As these models become more pervasive and their applications expand, understanding them and identifying their failure modes becomes increasingly important.


\section{Conclusion}
\label{section:conclusion}

Explainable AI/ML is an emerging field with important implications for machine learning as a whole. With the increasing use of machine learning models in real-world applications, it's more important than ever that we understand such models and what they learn. Evolutionary computing is well-poised to contribute to the field, bringing a rich toolbox of tools for performing black-box optimization. In this chapter, we introduced various paradigms for explaining a machine learning model and current methods of doing so. We then discussed how evolutionary computing can fit into these paradigms and the advantages of employing it. In particular, evolutionary computing as an optimizer is well suited for tricky interpretability metrics which are difficult to handle due to reasons such as non-differentiability, as well as for population-based metrics such as diversity and for optimizing a multiple of these metrics at the same time. We highlighted a few methods in each category which leveraged some of these strengths, but there is still significant room for more exploration and more advanced evolutionary algorithms. 

The field of explainable machine learning is still new, and much knowledge remains locked away within trained models that we still do not have the means to decipher. The use of evolutionary computing for XAI is still uncommon, but there are many opportunities ripe for the picking and we believe that it has the potential to play a key part in the future of XAI.

\bibliographystyle{plain}
\bibliography{refs.bib}

\end{document}